# Real-Time Hand Gesture Recognition for OpenXR Using Transformer-Based Machine Learning

Salar Rezayani[1] Dr. Russell Butler[2]
[1]University of Waterloo [2]Bishop's University
srezayan@uwaterloo.ca · russell.butler@ubishops.ca

## Abstract

Hand gesture recognition is a key component in human-computer interaction (HCI), enabling intuitive interfaces for applications in gaming, virtual reality (VR), robotics, and more. This study integrates transformer-based machine-learning models for real-time hand gesture recognition, using hand-tracking data captured through the OpenXR standard in Unity. We leverage positional data of hand joints and wrist rotation angles to train a custom gesture recognition system. By utilizing the sequential modeling capabilities of transformers, the system captures temporal dependencies within short gesture windows and classifies gestures robustly across hand orientations and sizes. The results show a significant improvement in gesture classification accuracy. Building on this, we outline how the approach can be extended toward detecting the flow of movement — the transitions between gestures — as future work.

## 1. Introduction

The interaction between humans and machines has evolved significantly, with gestures becoming an intuitive medium of communication. Hand gesture recognition systems rely on tracking joint positions and wrist orientations to classify static gestures. However, detecting the flow of movement, which includes gesture transitions and continuous motion, remains a challenging task. Transformer models, known for their success in natural language processing and time-series analysis, offer a powerful approach for capturing temporal dependencies in gesture sequences. This paper proposes a transformer-based method for robust real-time gesture recognition from OpenXR hand-tracking data, and outlines how it can be extended toward flow-of-movement detection.

## 2. Related Work

Traditional gesture recognition systems primarily employ convolutional neural networks (CNNs) or recurrent neural networks (RNNs) for static and dynamic gestures. While CNNs excel at extracting spatial features, RNNs focus on sequential dependencies. However, RNNs suffer from limitations such as vanishing gradients and difficulty capturing long-term dependencies. Transformers, with their attention mechanisms, address these issues by learning global relationships within sequences, motivating their use for hand gesture recognition and, in future work, for detecting the flow of movement between gestures.

Several recent works apply transformers to hand gesture recognition. The first group uses image- or signal-based inputs, each different from the tracked joint coordinates used here:

- *GestFormer: Multiscale Wavelet Pooling Transformer Network for Dynamic Hand Gesture Recognition* (Garg et al., CVPRW 2024) proposes a resource-efficient transformer for dynamic hand gesture recognition. It replaces quadratic self-attention with a pooling-based token mixer (PoolFormer) and combines wavelet-transform features, multiscale pooling, and a gated mechanism. It is evaluated on the NVIDIA Dynamic Hand Gesture and Briareo datasets using multimodal video inputs (color, depth, infrared, surface normals, and optical flow), rather than tracked 3D joint coordinates as in our work.

- *TraHGR: Transformer for Hand Gesture Recognition via Electromyography* (Zabihi et al., IEEE TNSRE 2023) addresses gesture recognition from surface electromyography (sEMG) signals for myoelectric prosthesis control. It uses a hybrid transformer with two parallel paths fused by a linear layer and is evaluated on the Ninapro DB2 dataset (40 users, 49 gestures). The sensing modality is muscle activity rather than spatial hand pose, but it demonstrates the strength of transformers for gesture classification.
- *A Transformer-Based Network for Dynamic Hand Gesture Recognition* (D'Eusanio et al., 3DV 2020) is the most closely related in task: it applies a transformer to dynamic hand gesture recognition from depth maps and surface normals captured by a single depth sensor, reaching state-of-the-art results on the NVIDIA Dynamic Hand Gesture and Briareo (automotive) datasets. It relies on image-based depth input rather than the tracked joint positions and wrist rotations used here.

Most relevant to our setup are skeleton-based methods, which operate directly on 3D hand-joint coordinates, as we do:

- *Construct Dynamic Graphs for Hand Gesture Recognition via Spatial-Temporal Attention (DG-STA)* (Chen et al., BMVC 2019) builds a fully connected graph from the hand skeleton in which node features and edges are learned by a self-attention mechanism operating in both the spatial and temporal domains. Using only joint positions, it reaches strong results on the DHG-14/28 and SHREC'17 benchmarks, and a spatial-temporal mask reduces computational cost substantially. This is closely aligned with our use of 3D joint coordinates and attention over time.
- *Real-Time Monocular Skeleton-Based Hand Gesture Recognition Using 3D-Jointsformer* (Zhong et al., Sensors 2023) combines a 3D-CNN that computes high-level skeleton embeddings with a transformer that uses self-attention to capture long-range temporal dependencies in the joint sequence. It is designed for real-time recognition from estimated hand skeletons, a goal shared by our system.

Together, these works show that transformers are effective for hand gesture recognition across video, depth, EMG, and skeleton (3D-joint) inputs. Our contribution differs in operating on tracked 3D hand-joint positions and wrist rotations captured through OpenXR hand-tracking in Unity for real-time recognition, with flow-of-movement detection — the transitions between gestures — identified as the next step.

# 3. Methodology

## 3.1 Data Collection

Gesture data was collected using Unity's OpenXR framework, capturing the 3D positions of 21 hand joints and wrist rotation angles (pitch, yaw, and roll). Each joint was visually represented with a cube, color-coded for the local X (red), Y (green), and Z (blue) axes to provide clear rotational context. Data normalization ensured that the system accounted for varying hand sizes by referencing all measurements relative to the palm.

An open-hand gesture was used as the baseline reference, with joints labeled from the wrist to the palm and across each finger (index, middle, ring, little, and thumb). Distances between joints and rotations were key features captured to enhance the robustness of gesture recognition.

The dataset included diverse hand orientations and rotations, ensuring comprehensive coverage of all possible angles and directions. For visualization, see Figure 3.1, which depicts the ball-grabbing gesture with visualized joints and color-coded axes on each joint.

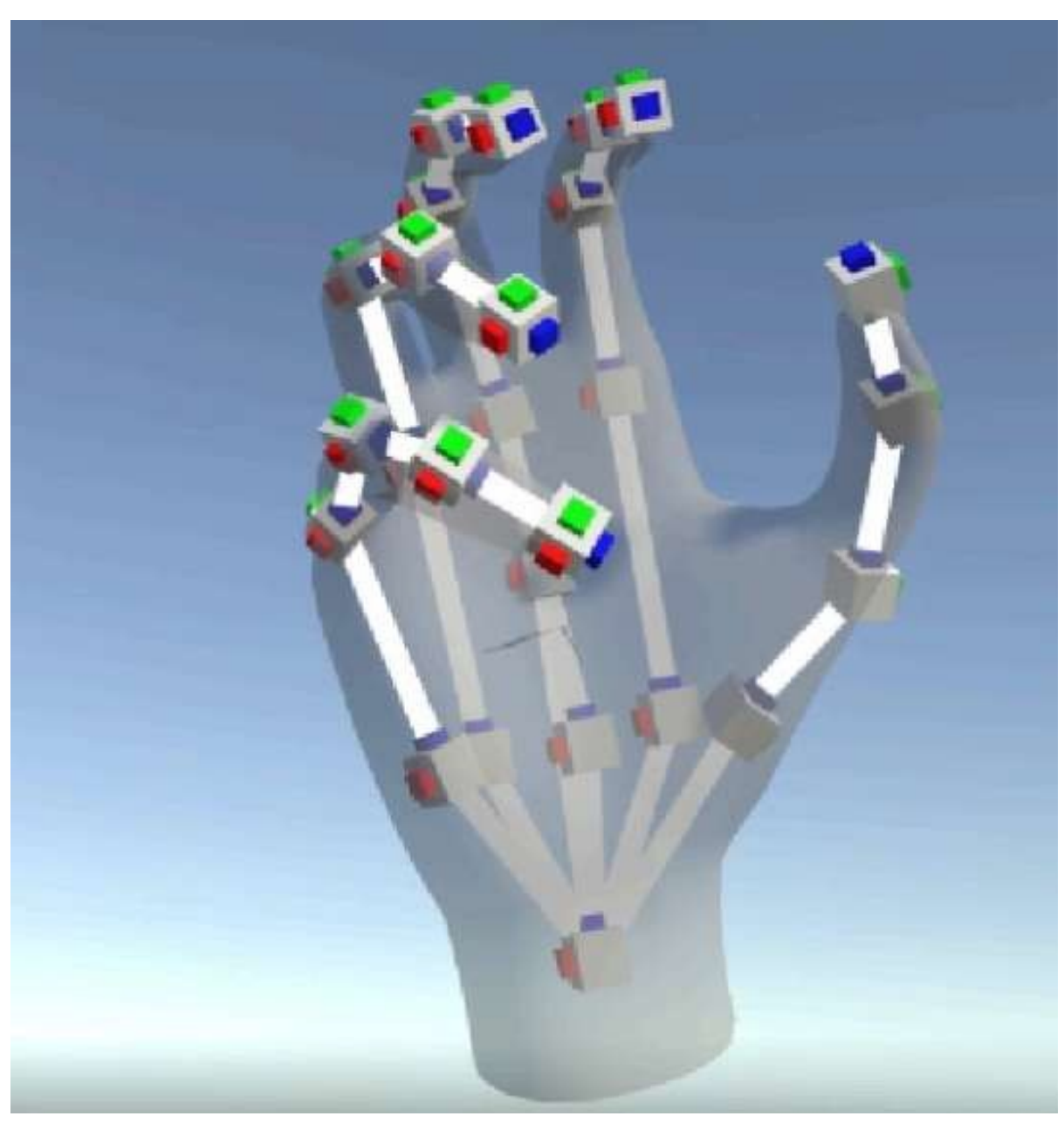

*Figure 3.1: The ball-grab gesture, showing visualized joints with color-coded local X (red), Y (green), and Z (blue) axes.*

## 3.2 Data Preprocessing

Data preprocessing involved normalizing joint positions relative to the palm and scaling by the wrist-to-palm distance. Wrist rotations were described using both numerical Euler angles and human-readable labels such as "Upward," "Turned Left," and "Neutral Roll." Gesture sequences were segmented into overlapping windows of 30 frames to capture short-term motion context.

## 3.3 Transformer Model Architecture

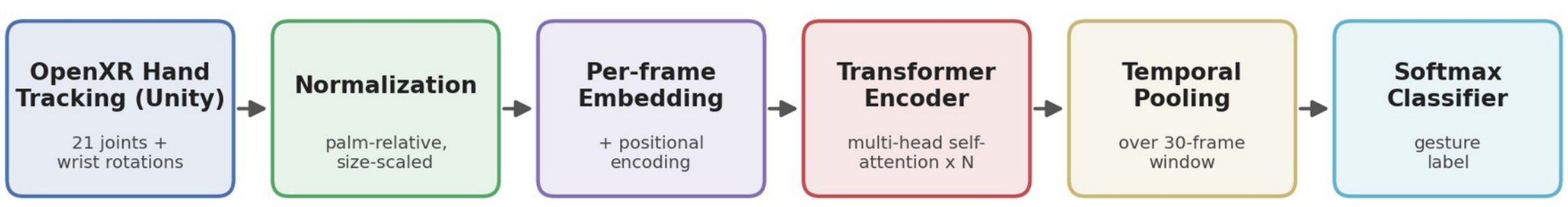

*Figure 3.2: System pipeline, from OpenXR hand-tracking input to gesture prediction.*

The transformer model was designed to process short sequences of hand-tracking frames and classify the gesture being performed. Key components included:

- **Input Embedding:** each time step's positional and rotational data was embedded into a high-dimensional space.
- **Positional Encoding:** added to the embeddings to retain temporal information.
- **Transformer Layers:** comprised multi-head self-attention mechanisms and feed-forward neural networks to capture dependencies between frames.
- **Output Layer:** a softmax classifier to predict the gesture label.

The attention mechanism in transformers calculates the relationship between different time steps using the following formula:

$$Attention(Q, K, V) = softmax(QK^T / \sqrt{d_k})\, V$$

Where:

- $Q$: query matrix representing the current frame.
- $K$: key matrix representing all frames.
- $V$: value matrix containing feature representations of all frames.
- $d_k$: dimensionality of the key vectors.

The attention mechanism ensures the model focuses on the frames within the window that are most relevant for recognizing the gesture.

### 3.4 Training and Evaluation

The model was trained using the Adam optimizer with a categorical cross-entropy loss:

$$L(y, \hat{y}) = -(1/N) \Sigma_{i=1}^{N} \Sigma_{j=1}^{c} y_{ij} \log(\hat{y}_{ij})$$

Where:

- $y_{ij}$: ground-truth label for class j of sample i.
- $\hat{y}_{ij}$: predicted probability for class j of sample i.
- $N$: total number of samples.
- $C$: total number of classes.

The dataset was split into 80% training and 20% testing sets, ensuring balanced representation of gestures. Performance was measured using classification accuracy and F1-score over the gesture classes.

## 4. Results

### 4.1 Static Gesture Recognition

The transformer model achieved a classification accuracy of 93.7% on static gestures, outperforming baseline RNN and CNN models. Notably, the model demonstrated robustness in differentiating gestures that were similar in shape and orientation. For instance:

- Gestures such as “Point” and “Gun,” often challenging due to their similar finger positions, were consistently predicted correctly.
- Slight variations, such as opening or closing individual fingers in “Like,” still led to accurate predictions until a threshold was crossed (e.g., transitioning to “OK” when three fingers were opened).

## 4.2 Wrist Orientation

By incorporating wrist rotation angles, the system enhances the contextual understanding of gestures. For example, “Point” gestures with an “Upward” orientation are correctly differentiated from those with a “Neutral Roll.” To achieve consistent recognition, the data is normalized and localized based on the player’s rotation, so that the player’s overall rotation does not affect the recognition process. We note that reliable wrist-orientation estimation is more difficult for some hand poses, and orientation-dependent gestures are therefore more sensitive to tracking noise than orientation-invariant ones.

## 4.3 Visual Results

Each gesture is visually represented with its corresponding predicted posture:

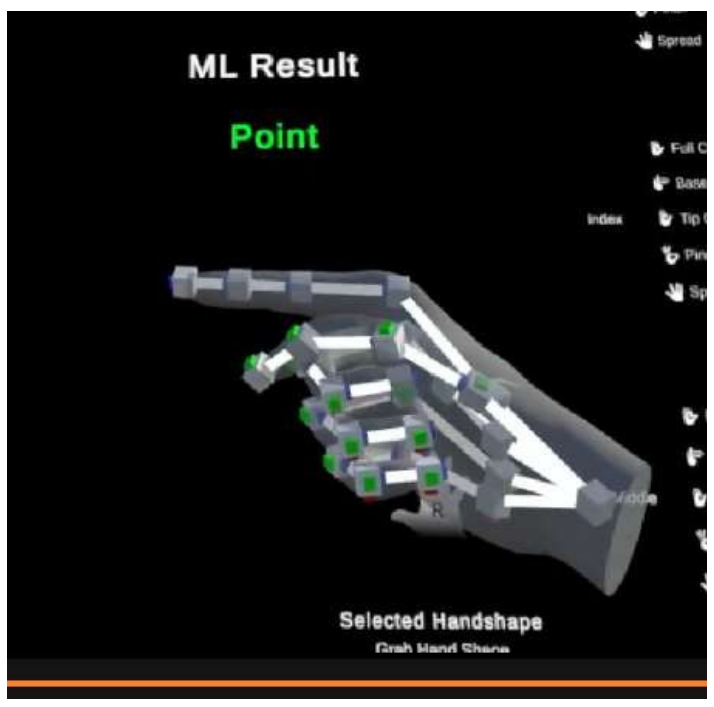

*Figure 4.1: Pointing gesture*

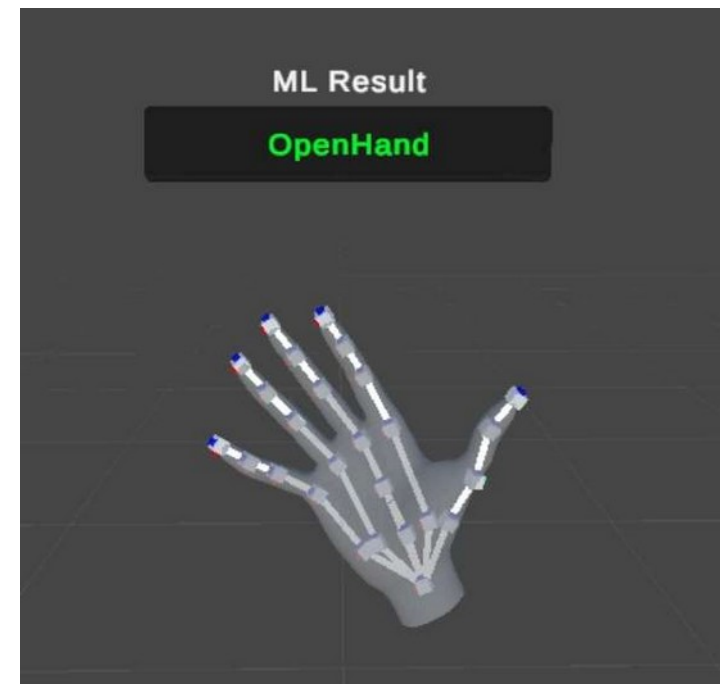

*Figure 4.2: Open Hand gesture*

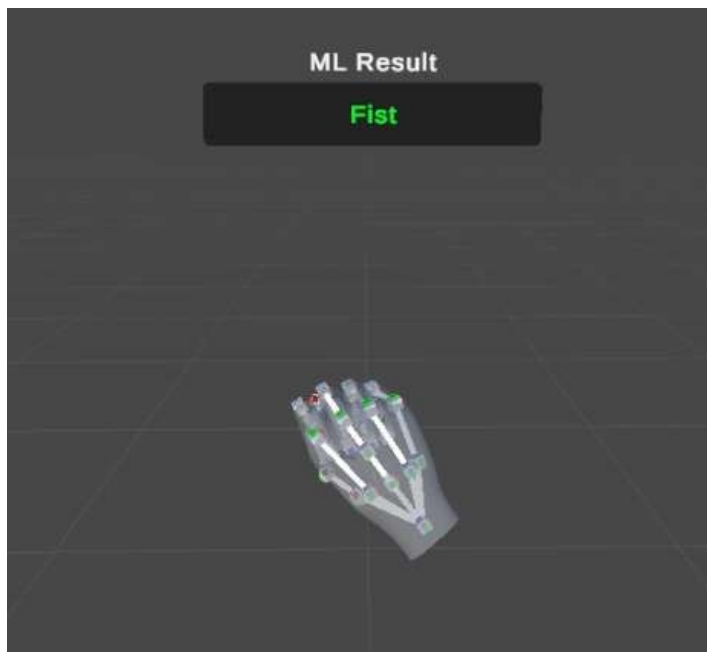

*Figure 4.3: Fist gesture*

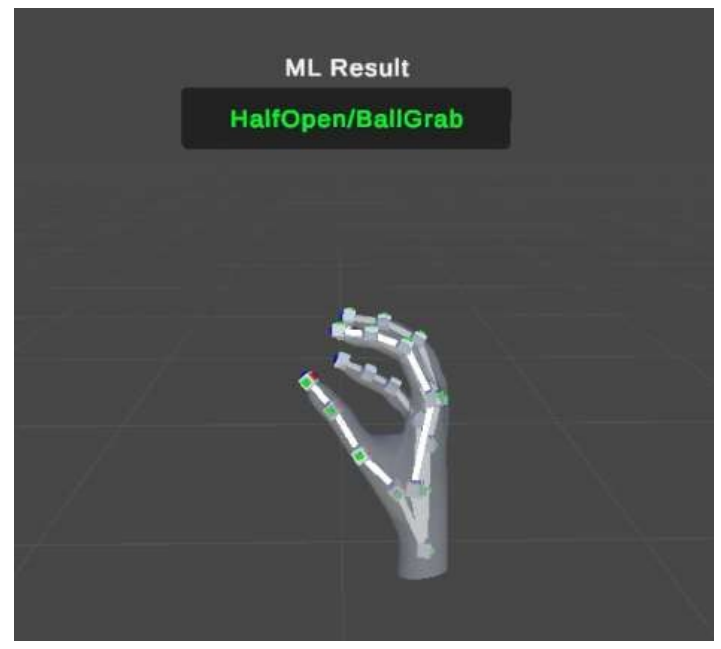

*Figure 4.4: Ball Grab gesture*

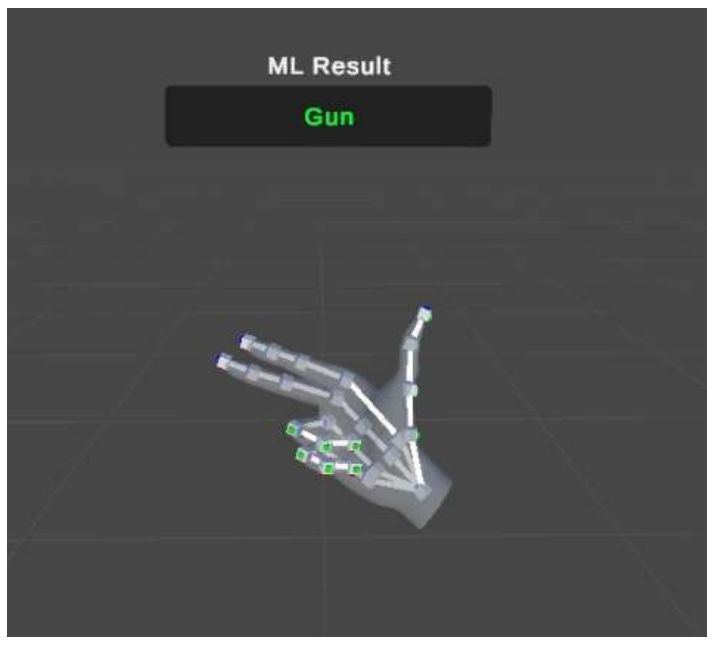

*Figure 4.5: Gun gesture*

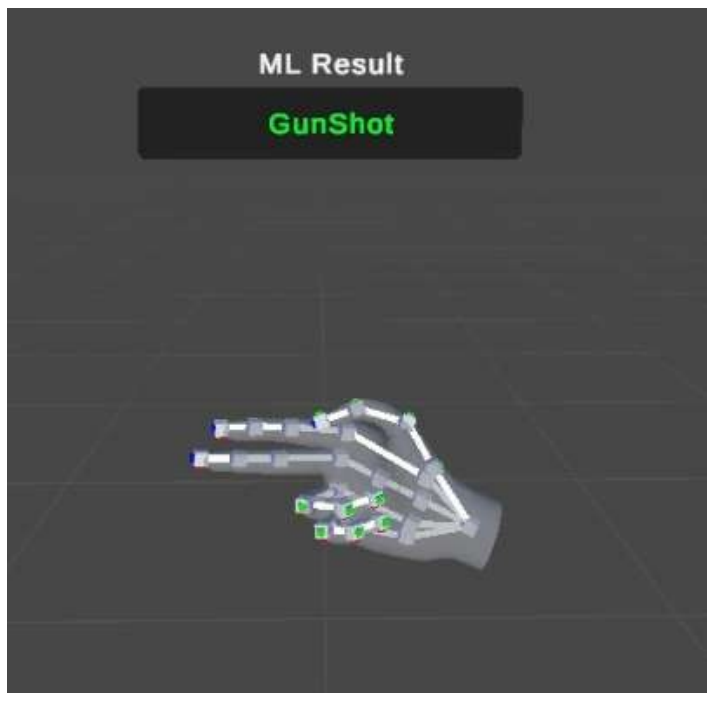

*Figure 4.6: Gun Shot gesture*

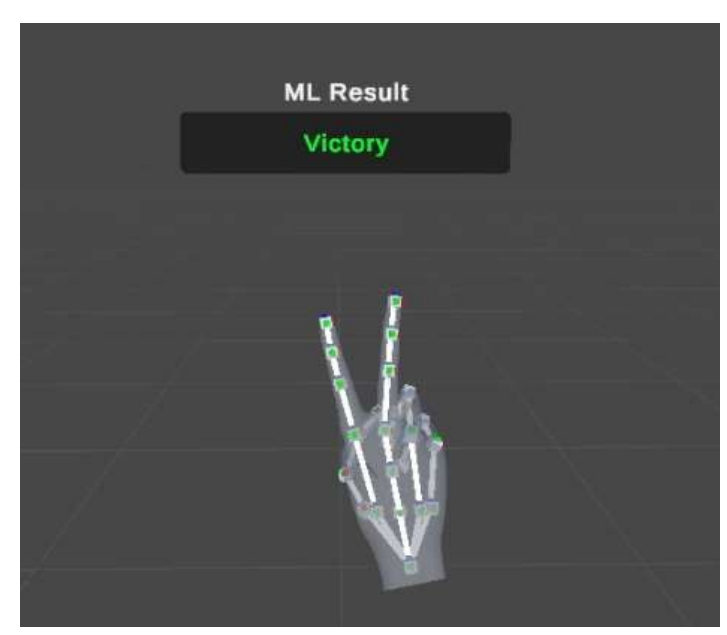

Figure 4.7: Victory gesture

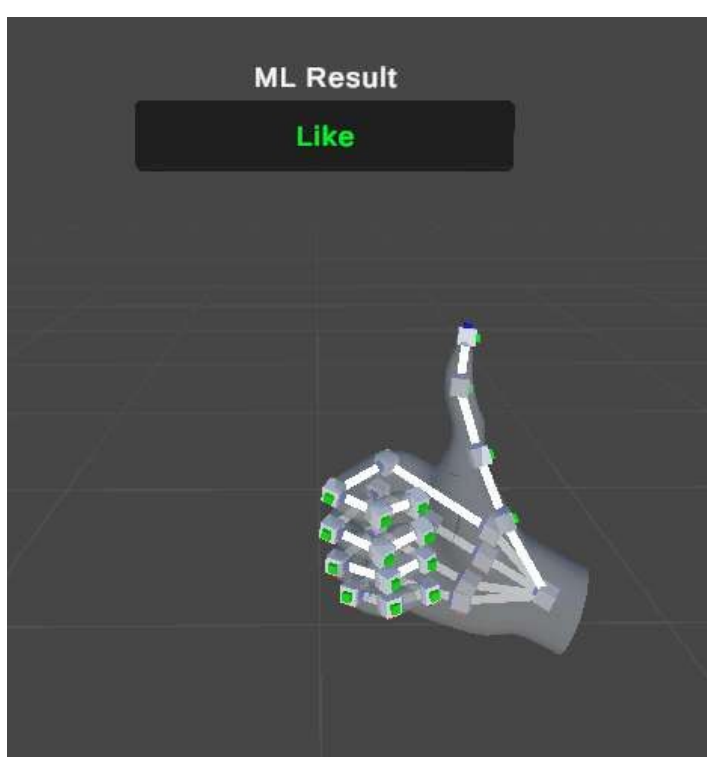

Figure 4.8: Like gesture

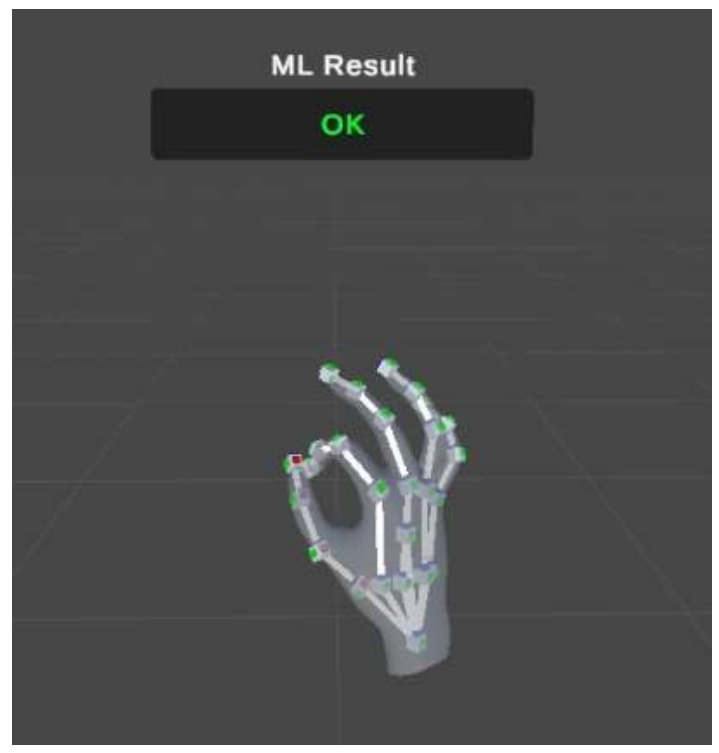

Figure 4.9: OK (Perfect) gesture

## 4.4 Comparison with Public Benchmarks

The results above were obtained on a custom dataset collected in Unity. To position this work within the broader field and to enable a direct, reproducible comparison with prior methods, this section summarizes the standard public benchmarks used by transformer- and skeleton-based gesture recognition systems and reports their published accuracies. Evaluating our model on these datasets, under the same protocols, is the natural next step and is required before any claim of state-of-the-art performance.

The most relevant public datasets are:

- **SHREC'17 Track:** 2,800 dynamic gesture sequences performed by 28 subjects, evaluated in both 14-gesture and 28-gesture settings using 3D hand-skeleton joints. This is the closest match to our joint-based input.
- **DHG-14/28:** 14 and 28 dynamic gestures captured with depth and 22-joint skeleton data, evaluated with leave-one-subject-out cross-validation over 20 subjects.
- **NVGesture and Briareo:** multimodal (RGB, depth, infrared, surface-normal) dynamic gesture datasets used by video-based transformer methods such as GestFormer and D'Eusanio et al.

Table 2 reports the accuracy of representative skeleton-based methods on SHREC'17 and DHG-14/28, as compiled by Cui et al. (2025). These serve as the reference points our model should be measured against once it is evaluated under the same protocols.

| Method | Venue | SHREC-14 | SHREC-28 | DHG-14 | DHG-28 |
|---|---|---|---|---|---|
| ST-GCN | AAAI 2018 | 92.7 | 87.7 | 91.2 | 87.1 |
| Shift-GCN | CVPR 2020 | 95.5 | 89.4 | 93.2 | 87.4 |
| HPEV | CVPR 2020 | 94.9 | 92.3 | 92.5 | 88.9 |
| CTR-GCN | ICCV 2021 | 96.1 | 94.4 | 93.1 | 90.5 |
| TD-GCN | IEEE TMM 2023 | 97.02 | 95.36 | 93.9 | 91.4 |
| DSTSA-GCN | 2025 | 97.74 | 95.37 | 95.04 | 93.57 |
| *Our method (this work)* | — | — | — | — | — |

*Table 2: Reported recognition accuracy (%) of representative skeleton-based methods on the SHREC'17 Track and DHG-14/28 benchmarks (14- and 28-gesture settings). Values are taken from the original papers as compiled by Cui et al.*

*(2025). Our method has not yet been evaluated on these public datasets; the final row is a placeholder to be completed after running the standard protocols.*

To obtain comparable numbers, the model should be applied to SHREC'17 using the official train/test split (1,960 training and 840 test sequences), reporting both 14- and 28-class accuracy; and to DHG-14/28 using leave-one-subject-out cross-validation over the 20 subjects, reporting the average. This requires mapping our 21-joint OpenXR hand skeleton to each dataset's joint convention and retraining with matched input features (normalized joint positions and angles). Reporting accuracy under these standard protocols would allow a direct, fair comparison with the methods in Table 2.

## 5. Discussion

The integration of transformers in gesture recognition opens new possibilities for real-time applications. Modeling temporal dependencies within a short frame window, rather than classifying a single frame, makes recognition more stable, and the same sequential capacity is a natural foundation for future flow-of-movement detection — recognizing the transitions between gestures — which is a direction we have not yet evaluated. The inclusion of wrist orientation adds an additional layer of context, making the system more robust for diverse applications, though reliable orientation is harder to obtain for some poses. Limitations include the modest size of the custom dataset and the absence of public-benchmark evaluation.

This study highlights the effectiveness of transformers for real-time hand gesture recognition. The inclusion of comprehensive data, normalization based on the palm, and attention mechanisms significantly improved accuracy. Observations include:

1. **Handling close gestures:** gestures like "Like" and "OK," as well as "Gun" and "Point," were accurately predicted thanks to attention mechanisms and contextual wrist rotation.
2. **Robustness across rotations:** normalization ensured gestures were recognized consistently regardless of hand orientation, making the model adaptable to diverse use cases. For some gestures, such as "Victory" or "Like," it is important to predict correctly when the hand is upward; for others, such as "Flat," "Gun," and "Gunshot," orientation does not matter.

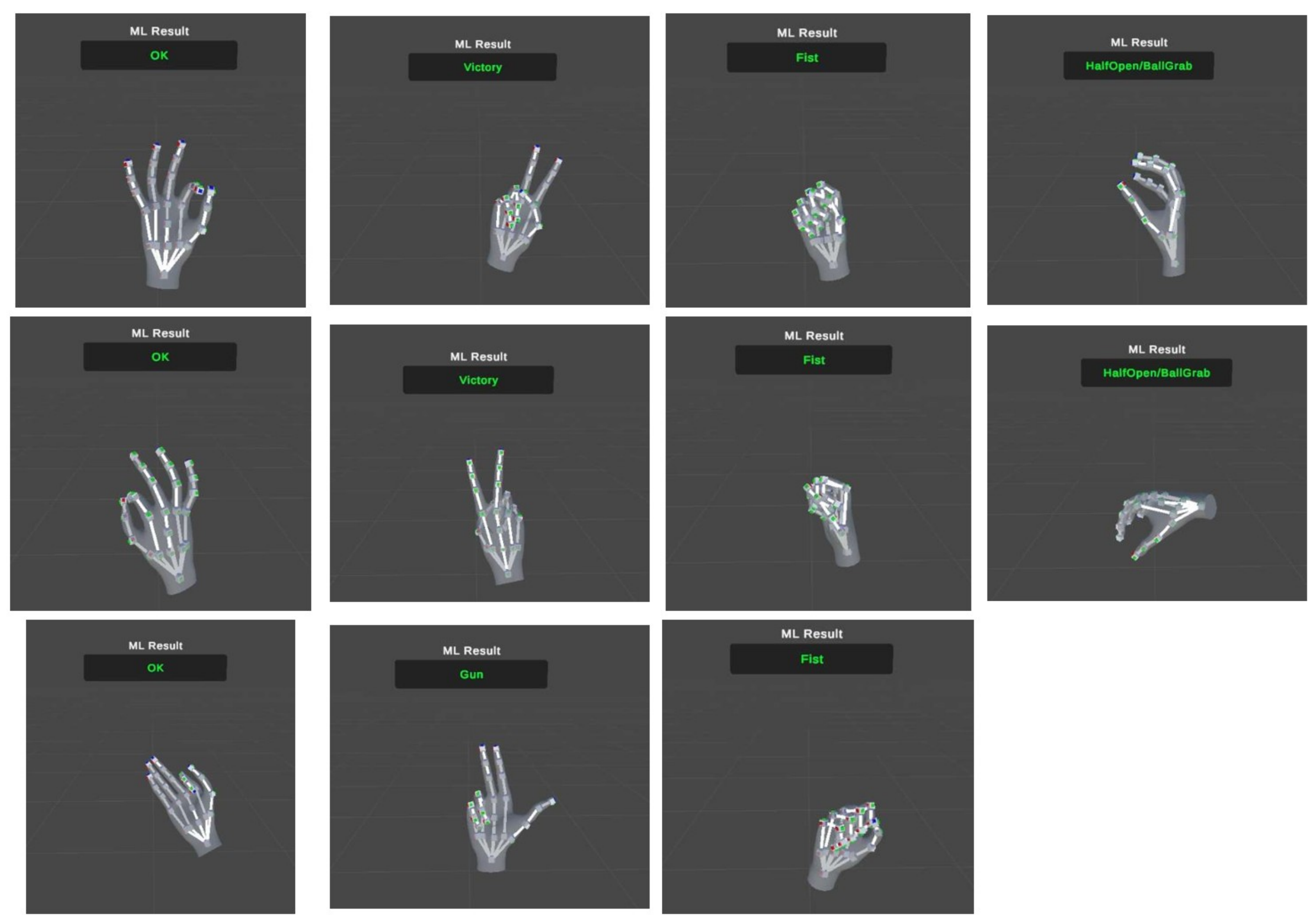

*Figure 5.1: The rotation independency for specific gestures. The same gesture is recognized correctly across a range of hand orientations.*

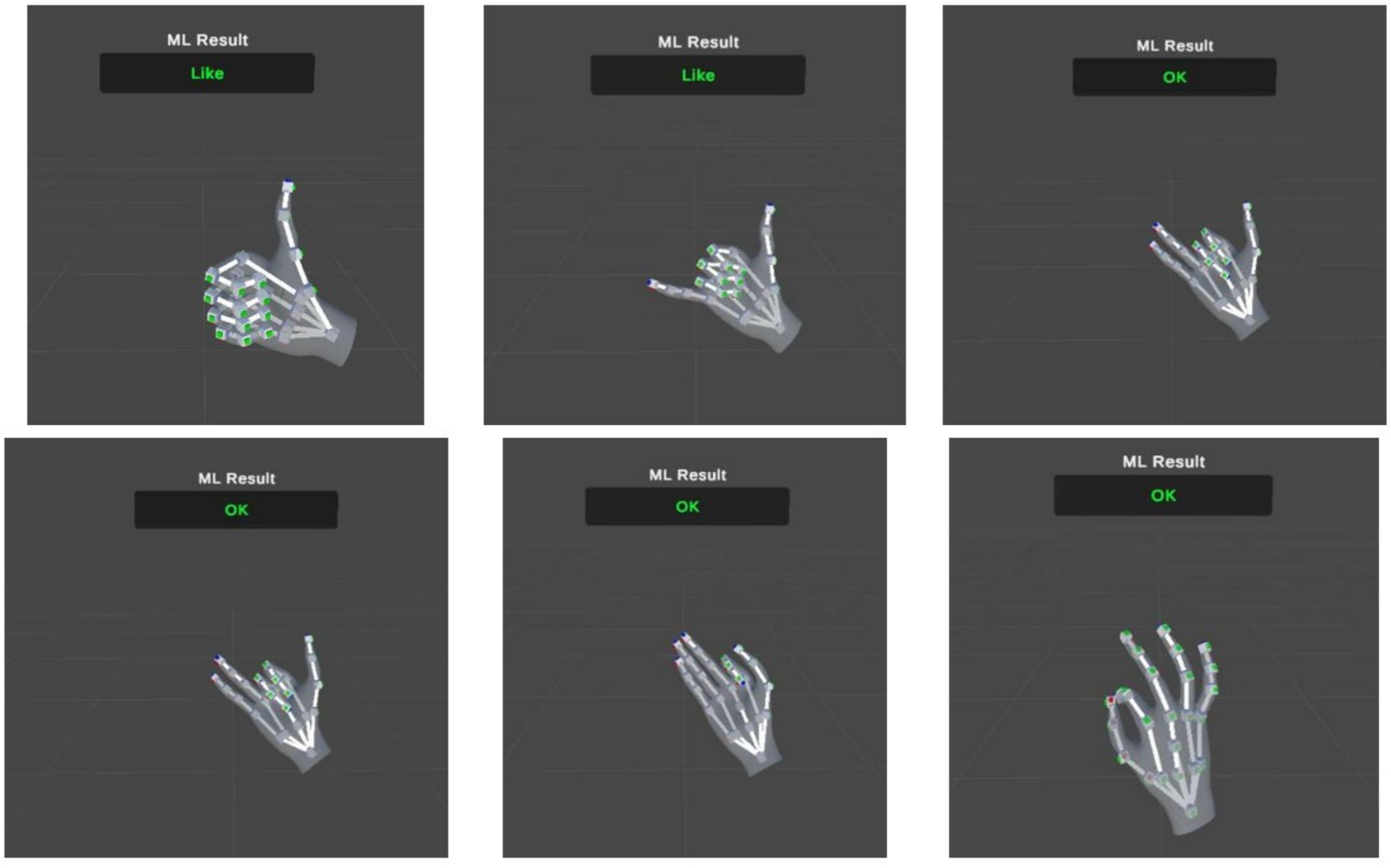

*Figure 5.2: Gesture rotation and similarity prediction.*

Figure 5.2 illustrates gesture rotation and similarity prediction. The model predicts gestures based on their closest similar shape. For example, in a “Like” gesture, if the pinky finger is extended, the model still predicts “Like.” This also applies to the second finger. However, if the third finger is extended instead, the model predicts “OK” (Perfect).

# 6. Future Work

Future research will focus on:

3. **Real-time optimization:** adapting the model for deployment on low-latency systems in Unity using ONNX and Barracuda.
4. **Flow-of-movement detection:** extending the model from classifying isolated gestures to detecting the transitions between gestures and continuous motion, such as sign language. This is the primary direction of ongoing work and is not yet evaluated in this paper.
5. **Multimodal inputs:** incorporating additional data such as hand velocity and acceleration for richer motion analysis.

# 7. Conclusion

This paper demonstrates the effectiveness of transformer models for real-time hand gesture recognition from OpenXR hand-tracking data. By leveraging temporal dependencies within short windows and integrating wrist orientation, the system achieves high accuracy and contextual understanding, and provides a foundation for future flow-of-movement detection. The proposed method is a step toward more intuitive and dynamic gesture-based interfaces, paving the way for advanced HCI systems in gaming, VR, and beyond.

# Appendix: Impact of Data Collection Refinements

## A.1 Data Collection Challenges

In the initial data collection process, the dataset lacked sufficient coverage of hand rotations, directions, and diverse hand sizes. This led to suboptimal model performance due to underrepresentation of critical poses and orientations.

## A.2 Normalization and Comprehensive Data Gathering

To address these issues, the data collection process was refined to include:

- Full coverage of hand rotations and orientations (pitch, yaw, roll).
- Normalization of joint positions relative to the palm, accounting for varying hand sizes.
- Inclusion of all possible distances between joints.

## A.3 Comparative Results

The table below compares gesture recognition accuracy before and after the improved data collection:

| Gesture | Accuracy Before (%) | Accuracy After (%) |
|---|---|---|
| Point | 61.47 | 94.18 |
| Fist | 36.77 | 91.82 |
| OK | 19.95 | 99.42 |
| Half Open or Ball Grab | 21.9 | 62.05 |
| Victory | 42.44 | 90.79 |
| GunShot | 2.84 | 22.94 |
| OpenHand | 22.92 | 97.77 |
| Like | 24.03 | 82.86 |
| Gun | 12.8 | 80.79 |

*Table A.3.1: Gesture recognition accuracy before and after the improved data collection.*

### A.4 Visualization

The impact of data collection refinements is visualized in the figure below:

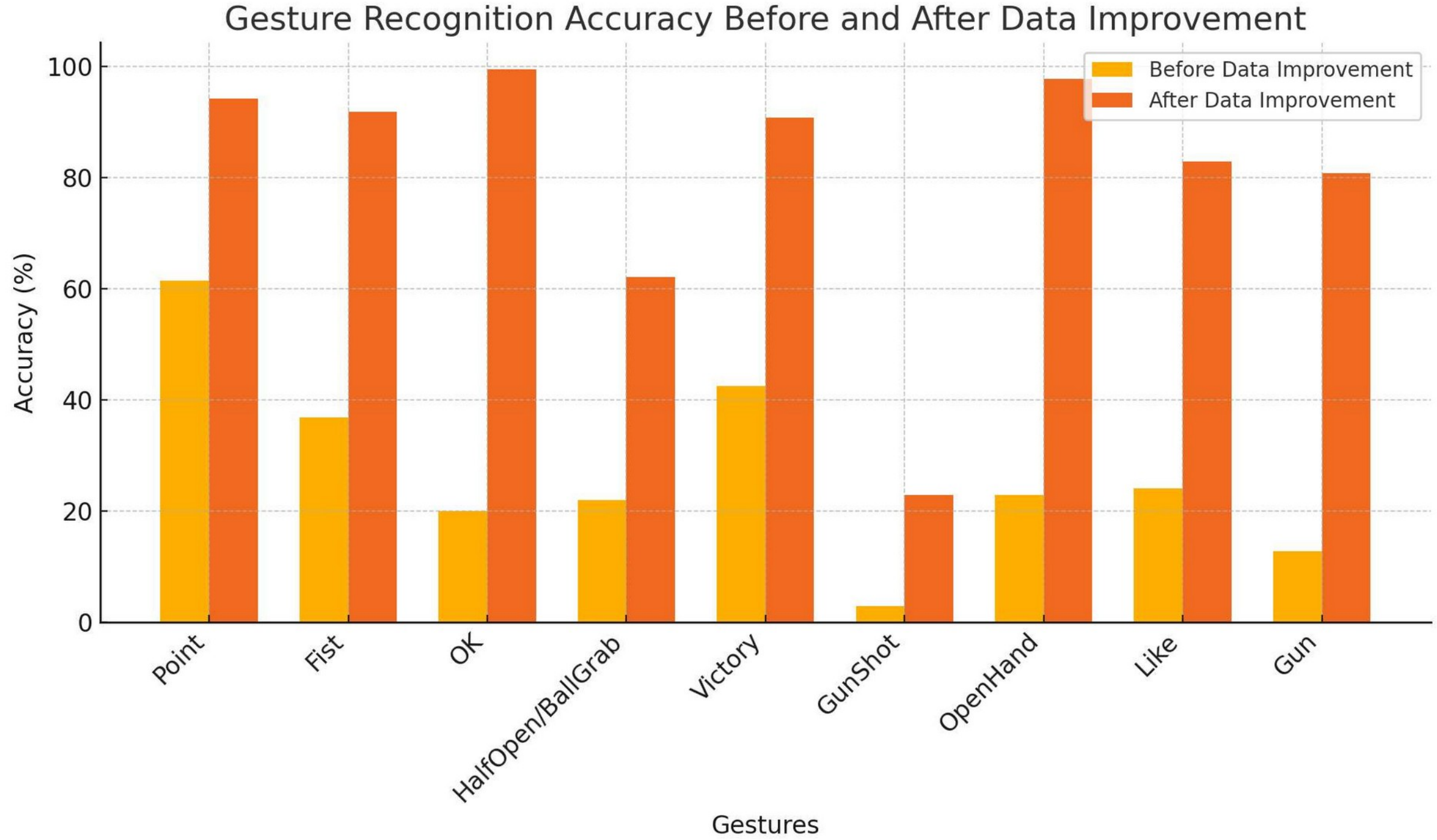

*Figure A.1: Gesture recognition accuracy before and after improved data collection.*

## Declarations

**Ethics and consent.** The hand-tracking data used in this study was collected by the authors and consists solely of hand-joint positions and wrist rotations captured in Unity via OpenXR; it contains no images, faces, or other personally identifiable information. As the data comprises only non-identifying hand-motion recordings gathered by the researchers, no separate ethics-board review was sought.

**Use of AI tools.** AI-based tools were used to assist with drafting and editing portions of this manuscript, including language, figures, and reference formatting. All technical content, system design, data collection, experiments, and results are the authors' own work, and the authors have reviewed and verified the final text, figures, and citations.